# Parkinson's Disease Assessment from a Wrist-Worn Wearable Sensor in Free-Living Conditions: Deep Ensemble Learning and Visualization

Terry Taewoong Um · Franz Michael Josef Pfister · Daniel Christian Pichler · Satoshi Endo · Muriel Lang · Sandra Hirche · Urban Fietzek · Dana Kulić

**Abstract** Parkinson's Disease (PD) is characterized by disorders in motor function such as freezing of gait, rest tremor, rigidity, and slowed and hyposcaled movements. Medication with dopaminergic medication may alleviate those motor symptoms, however, side-effects may include uncontrolled movements, known as dyskinesia. In this paper, an automatic PD motor-state assessment in free-living conditions is proposed using an accelerometer in a wrist-worn wearable sensor. In particular, an ensemble of convolutional neural networks (CNNs) is applied to capture the large variability of daily-living activities and overcome the dissimilarity between training and test patients due to the inter-patient variability. In addition, class activation map (CAM), a visualization technique for CNNs, is applied for providing an interpretation of the results.

Terry Taewoong Um and Dana Kulić

Department of Electrical and Computer Engineering, University of Waterloo, Waterloo, ON, Canada

E-mail: terry.t.um@gmail.com, dana.kulic@uwaterloo.ca

Franz Michael Josef Pfister

Department of Informatics and Statistics, Ludwig-Maximilians University, Munich, Germany

E-mail: fmj.pfister@me.com

Daniel Christian Pichler

Department of Neurology, Technical University of Munich, Germany

E-mail: daniel.pichler@tum.de

Satoshi Endo, Muriel Lang, and Sandra Hirche

Department of Electrical and Computer Engineering, Technical University of Munich, Germany

E-mail: {s.endo, muriel.lang, hirche}@tum.de

Urban Fietzek

Department of Neurology and Clinical Neurophysiology, Schön Klinik München Schwabing, Germany

E-mail: urban.fietzek@schoen-kliniken.de

# 1 Introduction

## 1.1 Parkinson's disease assessment using wearable sensors

Most Parkinson's disease (PD) patients suffer from disorders in motor function known as Parkinson's syndrome, for example, stiff muscles, freezing of gait, rest tremor, and slowed movements. These PD symptoms can be alleviated by intake of dopaminergic medication. However, long-term use coupled with overdosing may induce uncontrollable fluctuating movements, called drug-induced dyskinesia[21]. A PD patient's motor state is classified as: *OFF* with Parkinson's syndrome symptoms, *DYS* with dyskinetic symptoms, and *ON* with no salient Parkinson's syndrome or dyskinetic symptoms observed[21].

It is important to accurately assess the PD patient's motor state for ensuring appropriate medication dosage. Assessing the motor state of a PD patient is usually performed in a clinical environment. Typically, the patient is asked to perform a series of specific test movements while under observation by a clinical expert[9]. As a result, PD assessment only occurs during clinical visits, and may not accurately capture symptoms at home and throughout the day. Moreover, the assessment by a clinician can be susceptible to inter-rater variability[22]. For these reasons, automatic PD assessment using wearable sensors has been actively investigated in the past decade[4,20].

In the home setting, PD motor symptoms usually co-occur with various unscripted movements. In addition, as there is no expert clinical observer, typically patients or their care-givers describe the motor state using diaries, but they are prone to mistakes[1]. As a result, PD assessment in free-living conditions is far more challenging than in controlled conditions. A large-capacity model, e.g., deep learning (DL)[17], could be used to capture the large variability of free-living movements. However, collecting a large-scale PD dataset from patients' daily life and having them labeled by clinical experts is impractical. Therefore, development of an automatic PD assessment suitable for free-living conditions remains an open problem[3].

Um et al.[24] approached the problem of limited data availability for DL by proposing data augmentation techniques, However, their success was limited to the classification of the two distinctive PD states, *OFF* and *DYS*, not including the *ON* state. In this paper, we extend the previous work to classify three PD states - *OFF*, *ON*, and *DYS* – by an ensemble of convolutional neural networks (CNNs)[16].

[Fig. 1 about here.]

Since a variety of voluntary activities can be observed in the asymptomatic state, the intra-class variability of the wearable-sensor data in *ON* state is large. Various voluntary movements in *ON* state can generate similar wearable-sensor signals to those in other states, e.g. fast voluntary movement that mimics the DYS state. In addition, the wearable-sensor may not be able to observe the motor symptom, e.g., tremor in the lower leg is not measured by the arm-worn sensor. These issues may lead to significant inter-class overlaps in the wearable-sensor data (Figure 1). Large variability between patients is another challenge that makes PD assessment more difficult by introducing dissimilar distributions between training and test data. Due to these increased challenges, PD assessment in the three PD state case is far more difficult than in the two PD state case.

We propose an ensemble of CNNs as a solution for this challenge. An ensemble model consisting of individual CNNs trained with different patient groups is applied to overcome the variability between patients. In addition, class activation map (CAM)[27] is applied to provide a visualization of the salient features of the signal used by the CNNs, enhancing the interpretability of the results.

## 1.2 Related Works

Most works for automatic PD assessment using wearable sensors, e.g., wrist-worn accelerometers, are based on small-scale data collected from a controlled environment during predefined motor tasks[4,20]. On the other hand, DL-based approaches present a promising avenue for PD assessment in natural environments[14] owing to the large expressive power and powerful data-driven feature identification of DL models. For instance, Hammerla et al.[10] classified four PD states, the three PD states plus the asleep state, using a restricted Boltzmann machine (RBM)[11]. However, the predictions were made hourly, which may be too coarse-grained for clinical purposes, e.g., for measuring the response of the levodopa therapy. Um et al.[24] successfully assessed PD states in free-living conditions at one-minute intervals. However, their success was limited to the two PD states, *OFF* and *DYS*, which is also not sufficient for clinical purposes. In short, PD assessment in free-living conditions remains an open problem, due to the challenges induced from the limited availability and large variability of the PD data[3,14].

DL is likely to produce unreliable predictions when a sufficient amount of training data is not available. In this case, the information of the predictive uncertainty of DL[8] plays an important role for detecting unreliable predictions, especially in medical applications[18]. As the calculation of the exact predictive uncertainty is computationally expensive in DL, recent techniques approximate the predictive uncertainty by utilizing, for example, dropout[7]. Lakshminarayanan et al.[15] showed that a simple ensemble model can effectively approximate the predictive uncertainty of DL if the objective function obeys a proper scoring rule. In their work, multiple neural networks with different initializations served as individual models of an ensemble for approximating predictive uncertainties. However, the obtained predictive uncertainty is not directly used for improving the performance of the target task. Also, there is still an unexplored question regarding *what individual models should be aggregated* for improving the performance of the target task.

Diversity of the individual models is crucial for the performance of an ensemble model. In the extreme case, no additional benefits are obtained from the ensemble if the individual models are fully correlated. Thus, it is important to aggregate individual models that are diverse, but give reasonable predictions[28]. In this research, individual CNNs trained with different patient groups are aggregated to form an ensemble to overcome the inter-patient variability.

CNN visualization is another avenue to improve the usability of CNN-based PD assessment. Various visualization techniques have been proposed for giving interpretable outputs from CNNs, e.g., saliency maps[26] and activation maps[23]. A class activation map (CAM)[27] is a simple but powerful approach to highlight the attended parts of the input without any localization labels. For an in-depth analysis of the results, wearable-sensor data are interpreted using a CAM in this research.

## 2 Materials and Methods

### 2.1 PD Dataset

Data were collected from 30 PD patients using a wrist-worn wearable sensor, the Microsoft Band 2. (The study was approved by the ethics committee of Technical University of Munich (Az. 234/16 S) on June 30, 2016.) The 30 PD patients are $67\pm10$ years old, median Hoehn & Yahr stage 3, average disease duration $11\pm5$ years, and MoCA points $26\pm3$. Microsoft Band 2 measures the three-dimensional acceleration and angular velocity of the arm movements with a frequency of 62.5Hz. Note that only the three Cartesian dimensions of acceleration data are used for this research as in most previous work[5,10,24]. During preprocessing, irregular timesteps of the raw data are corrected by resampling at a frequency of 60Hz. As a result, each one-minute window of data is a 3600×3 matrix, which will be regarded as a rectangular image input for the CNN[24,25].

[Fig. 2 about here.]

The wearable-sensor data are collected in free-living conditions from morning to evening. To collect and label the data, a clinical expert shadowed the 30 PD patients and rated their PD symptoms every minute. Based on the observed symptoms and their severities, patients' PD motor states are labeled into four levels of *OFF* and *DYS* plus a single *ON* state (Figure 2). In this research, the four severity levels are neglected and only three classes are considered - *OFF*: where Parkinson's syndrome symptoms are observed, *DYS*: where dyskinetic symptoms are observed, and *ON*: where neither Parkinson's syndrome nor dyskinetic symptoms are observed. In total, 11567-min data are collected from the 30 PD patients, of which 10977-min data are labeled into the three PD states. The remaining 590-min data are unlabeled and consist of private activities (e.g. taking shower).

Within the 10977 min of data, there exist a significant amount of *no-motion* data with different labels (Figure 2). There could be several reasons for the no-motion data: during the one minute, a patient in the OFF state may have experienced severely diminished arm motion, or a patient in the ON state may have been taking a rest, or a patient in the DYS state may have been suppressing the symptoms by, e.g., holding a fixed support. Also, no-motion data can be generated during the periods when the patient takes off the wearable device or the device has a problem in communication.

Since no-motion data do not present evidence for assessing the patient's PD states, it is reasonable to exclude them during training and predict them as "*no-motion*" at runtime. For this reason, 2316 instances of 1-min data, which have an average variance of acceleration magnitude of less than $2.75 \times 10^{-4} G^2$ within the 1-min window, are removed from the 10977-min data. The remaining 8661-min data are used for classifier training and evaluation, as described in the next section.

[Table 1 about here.]

Different PD patients have different PD distributions (Table 1). For example, Subj #0, #19, #21 have *OFF*-state-majority distributions while Subj #7, #17, #22 *DYS*-state-majority distributions. Note that most patients have *ON* state, thus, the training set collected from multiple patients is likely to have an *ON*-state-majority distribution. As a result, a fixed training set often fails to model the differing PD distributions of the target patients.

In Section 2.3, an ensemble of CNNs is introduced to overcome the challenge of inter-patient variability.

## 2.2 CNN architecture

A 7-layer CNN is employed for classifying the three PD states. To reduce the number of parameters and prevent overfitting, we replace fully-connected layers with a global average pooling (GAP) layer[19]. The GAP layer also enables the visualization using CAM, which will be discussed in Section 4.4.

[Fig. 3 about here.]

Figure 3 depicts the architecture of the 7-layer CNN. Each CNN layer consists of convolution (Conv), batch-normalization (BN)[12], and rectification (ReLU) layers and forms 64-128-256-512-1024-1024-1024 feature maps. For abstracting the feature maps through layers, strided convolutions are applied instead of using pooling layers. The size of the $3600 \times 3$ dimensional inputs reduces to $27 \times 1$ going through the convolution layers. The final 64 feature maps are averaged by the GAP layer, and finally, give three output-node values with a linear layer and *softmax* activation function.

## 2.3 Deep Ensemble Models

To overcome the problem of inter-patient-data variability, we build an ensemble classifier. The component models of the ensemble are trained with different training sets, in particular, with different patient-group data. The rationale behind the proposed ensemble model is that multiple models trained with different PD distributions may outperform a single model trained only with an *ON*-state-majority distribution. This rationale is similar to the principle of *sub-bagging*[2,6], which aims to obtain a stable learner by training on subsets of the data, i.e. subset of patients in this research.

To generate the ensembles, data from 15 randomly-selected patients are used for training an individual model of the ensemble. 100 models are trained and leave-one-subject-out is used for validation. In each fold, those models that do not include the test patient in their training sets are used to form the ensemble. In each fold, the ensemble thus consists of 46-54 individual models trained with different 15-patient groups. Note that these individual models necessarily have different PD distributions where some may be similar while others may differ from the target patient's distribution.

## 2.4 Aggregation over individual models

The ensemble model gives multiple predictions, one from each individual model. The remaining question is *how to combine these predictions to obtain a correct prediction*.

Majority voting is a straightforward way to aggregate multiple predictions. Note that the accuracy of the final answer asymptotically approaches 100% as more learners with at least 50% accuracy for each prediction are aggregated. Based on this fact, an ensemble model can aggregate predictions from weak learners and boost their performance using majority voting.

If the learners have different importance, their predictions can be aggregated with weights corresponding to their importance. Logit values can be used as the importance of each learner. Logits are the output node-values before softmax activation. If the input is easy to classify for a certain class, the node corresponding to the class will have high values while the other nodes will have low values. On the other hand, the output nodes will have similar values if the input is difficult to classify. For these reasons, logits are sometimes used for representing the uncertainty of the predictions, although this interpretation is theoretically not correct[8].

To incorporate the information from the logits, instead of aggregating the predictions (e.g. [0, 1, 0]) of all individual models, the logit values (e.g. [1.5, 2.7, 0.5]) from all individual models are summed and the final prediction is made based on the summed logits. This *logit-based weighted voting* assumes that the magnitude of logits has a weak correlation to the importance of each learner for the final prediction. Note that this weak correlation can be enhanced by aggregating multiple individual learners by the ensemble.

## 2.5 Aggregation over time

In addition to the aggregation across multiple learners, aggregation over time is also possible based on the assumption that nearby states are likely to be the same. In fact, 91.3% of consecutive three one-minute windows have the same PD state. Therefore, it is reasonable to pursue smooth predictions by aggregating predictions over time. Aggregation over time or prediction smoothing can also be interpreted as a simple temporal model on top of our instance-based learners. A heuristic that "*PD states change slowly*" can be a simple but powerful temporal model that does not require additional training data.

Note that aggregating over large neighborhoods may mask faster changing symptoms and introduces a lag during real-time prediction. Therefore, the optimal size of the smoothing kernel window can be different depending on the target patient, thus, has to be chosen by the experimenter as a hyperparameter. The effects of different kernel window sizes on PD assessment are evaluated in Section 3.5.

## 3 Results

### 3.1 Experiment setting

Experiments are performed using a computer with 4.00GHz×8 CPUs, 16GB RAM, and an NVIDIA GeForce GTX 1080 Ti GPU. Experimental code is written in Python 3.5 using the deep learning library Keras with TensorFlow backend. All models are trained for 40 epochs because all models achieved over 97% training accuracy and showed little progress after 40 epochs. Categorical cross-entropy is employed as the loss function and the ADAM optimizer[13] is applied for training with the default learning rate, 0.001. Note that no preprocessing, e.g. data augmentation[24], is applied to the raw data except for the resampling described in Section 2.1.

[Table 2 about here.]

### 3.2 Baseline CNN

A standard CNN is applied for obtaining a baseline for PD assessment. By treating each one-minute window as a rectangular image of 3600×3 dimensions, a CNN can straightforwardly classify the PD state similar to image classification[24,25]. In the experiments, one CNN is trained with 15 subjects (*CNN(n=15)*) and another CNN is trained with 29 subjects (*CNN(n=29)*), i.e. in a leave-one-out setting. Note that the results of the 15-subject CNN presented in Table 2 are the averaged results from 50 experiments with randomly selected 15 training subjects. As a result, testing accuracies of 50.97% and 54.16% are obtained from the 15-subject and 29-subject models, respectively.

[Table 3 about here.]

As another baseline, CNNs are trained with 10-fold cross-validation without distinguishing between patients. Since approximately 90% of the target patient's data are used for training, there is a smaller dissimilarity between the training and test datasets. As a result, an accuracy of 64.67%, which is 10.51% higher than *CNN(n=29)*, is achieved on average (Table 3).

### 3.3 Deep ensemble models

In the ensemble experiments, three different ensemble approaches are evaluated. First, dropout[7], which can be seen as an ensemble of differently-connected neural network models, is applied for generating multiple predictions (*EsbDropOut(n=29)*). From the 29-subject CNN model (*CNN(n=29)*), we randomly drop 30% of the connections and generate 50 predictions from 50 differently-connected or dropout models. A majority voting scheme is used for aggregating the 50 predictions and making the final prediction. As a result, an accuracy of 43.19%, which is the lowest performance among all experiments, is obtained.

The same models with different initializations[15] can also be used for forming an ensemble model (*EsbDiffInit(n=15)*). In these experiments, the 15-subject CNN model (*CNN(n=15)*) is initialized with 50 different parameter sets to form an ensemble model. As a result, the performance of the 15-subject CNN model (*CNN(n=15)*), 50.97%, is improved to 52.74% with a majority voting scheme, by mitigating the effect of initialization dependencies. Note that the results of *EsbDiffInit(n=15)* are the averaged results from 50 different models trained with 50 different 15-patient groups as in the *CNN(n=15)* experiment.

The final ensemble model that we propose is trained with 50 different patient groups where each group consists of 15 PD patients (*EsbDiffPat(n=15)*). Note that the proposed ensemble model, *EsbDiffPat(n=15)*, is different from *CNN(n=15)* or *EsbDiffInit(n=15)* in that it collects predictions from 50 different 15-patient groups and therefore exploits all 29 subject data instead of 15. In other words, *EsbDiffPat(n=15)* can be considered as a sub-bagging variation of CNN(n=29): the proposed ensemble model exploits all subject data, different initializations and a variety of PD symptom distributions from different patient groups. The proposed ensemble model achieves an accuracy of 59.14%, which is 4.98% and 8.17% higher than the baseline *CNN(n=29)* and *CNN(n=15)* models, respectively.

## 3.4 Aggregations over individual models

The results of *EsbDiffPat(n=15)* presented in Table 2 are based on a majority voting scheme, which assumes equal contributions of the predictions from the individual models. Instead, the predictions can be aggregated with different weights based on the logit or softmax values. Note that softmax-based voting takes into account the relative importance between nodes, but not between individual models.

In the results, majority, logit-based, and softmax-based voting schemes using the *EsbDiffPat(n=15)* ensembles obtain accuracies of 59.14%, 59.06%, and 59.04%, respectively. In fact, 93% and 97% of the final predictions given by the logit-based and the softmax-based voting schemes are identical to the predictions given by the majority voting scheme. In other words, the weights based on the logits or softmax values do not change a large portion of final predictions. Thus, a more aggressive voting strategy may be needed to change more final predictions and improve the performance over the majority voting scheme.

3.5 Aggregation over time

Based on the prior knowledge that PD states do not frequently change over time, predictions or logits can be aggregated not only over individual models, but also over neighboring minutes. Different kernel windows can be used for aggregating predictions over time. In the experiments, 0, 5, 11, 31, 61, 181, and infinite-minute kernel windows are applied for aggregating logit values. Note that the 0-minute-window results are identical to the results of the logit-based aggregation in the previous section while infinite-minute-window result is identical to giving predictions with the target patient's majority PD state.

[Table 4 about here.]

The results are presented in Table 4. The best performance is achieved by 11-minute uniform kernels with an accuracy of 63.05%, 3.99% higher than the no smoothing result. In fact, all prediction smoothing methods give similar performance from 61.60% to 63.05% regardless of kernel size. This result implies that prediction smoothing always gives similar additional performance improvement due to the outlier filtering effect.

## 4 Discussion

4.1 Ensemble

The respective baseline results with a single CNN, 50.97% and 54.16% for *CNN(n=15)* and *CNN(n=29)*, show that more data from additional patients is beneficial for CNN-based approaches. Also, the 10-fold cross-validation gives an accuracy of 64.67%, which is 10.51% higher than *CNN(n=29)*, despite the fact that it uses less data for training. One of the reasons for the high performance is that training data from the same target patient contributes to the prediction because they have similar data patterns and labeling conditions. In addition, the PD distributions of the training and the test datasets are similar in this case, thus, there may be fewer mispredictions due to the distribution dissimilarity.

The result of *CNN(n=15)* is improved by aggregating the predictions from the models with different initializations, *EsbDiffInit(n=15)*, from 50.97% to 52.74%. It implies that different initializations are one of the sources of prediction error. However, the performance of *EsbDiffInit(n=15)*, 52.74%, is lower than the 29-subject model's performance, 54.16%, most likely because it only exploits data of 15 patients rather than 29. In other words, using data from more patients provides more benefit than using variously initialized models.

Finally, the proposed ensemble model trained with different patient groups, *EsbDiffPat(n=15)*, achieved the best result, 59.14%, by exploiting the diversity of different patient groups. Interestingly, it presents a higher performance than *CNN(n=29)* which simply uses all data except for the target-patient data for training the model. It implies that our sub-bagging strategy with different patient groups is effective to overcome the large variability between patients by introducing diverse individual learners that reflect inter-patient variability.

Among the patients, Subj #10 and #9 show the largest performance improvements, 46.30% and 28.47%, respectively, between *EsbDiffPat(n=15)* and *CNN(n=29)*. Those patients have a large portion of DYS or ON: Subj #10 has 100% DYS-state data, and Subj #9 has 97% ON-state data. In other words, the proposed ensemble model aids with discriminating ON- and DYS-state data, which can look similar due to the effect of voluntary movements. On the other hand, *EsbDiffPat(n=15)* gives worse results than *CNN(n=29)*., e.g., for Subj #19, when individual learners provide too poor predictions to aggregate for a better final prediction.

## 4.2 Prediction with confidence information

[Fig. 4 about here.]

Figure 4(a) shows the performance according to the confidence values defined by the rankings based on voting agreements (blue), logit values (green), and softmax values (red). For example, the bars in the 0-20% section represent the accuracy of the predictions which rank in the top 20 based on voting agreement ratio, target logit values, and target softmax values. The results show that all three confidence estimates have strong correlations with their prediction accuracy: predictions with high confidence values give high-accuracy predictions while predictions with low confidence values give low-accuracy predictions. In particular, the predictions with the top 20% of logit-based confidence achieve an accuracy of 81.76%, which is 22.62% higher than the overall performance.

The PD distribution of the top 20% confidence data is different from the PD distribution of the whole dataset (Figure 4(b)). The whole dataset has 27%, 44%, and 29% of *OFF, ON,* and *DYS* state data, respectively, whereas, the top 20% confidence data has 39%, 28%, and 33%, respectively. On the other hand, a large amount of *ON*-state data can be observed in the uncertain predictions. This means that confident predictions can be made more easily by the CNNs for *OFF*-state data, while being more difficult for *ON*-state data. Since *ON*- and *DYS*-state data are more uncertain, their predictions can be altered through the aggregation process. This is the reason why our proposed ensemble achieves the largest performance improvement in *ON*- and *DYS*-state predictions.

[Table 5 about here.]

The top 20% confidence predictions are not evenly distributed over our 30 patients (Table 5): 66% of the data of Subj #21 belong to the top 20% confidence data whereas less than 1% of the data of Subj #14, #15, #28 belong to the top 20% confidence data. If the patient has dense high-confidence predictions, the entire-day prediction can be obtained by interpolating the high-confidence predictions. For example, an accuracy of 79.7% is achieved by interpolating the high-confidence data for the ten subjects for whom at least 30% of predictions belong to the top 20% confidence predictions.

4.3 Visualization

Although DL provides powerful performance when large-scale data are available, it is difficult to interpret the reason for its success or failure. Visualization techniques can enable clinicians to better understand and exploit the data-driven results.

Class activation map (CAM)[27] is a simple approach to visualize the parts of the input attended by a CNN. CAM automatically detects the salient parts of the input by measuring the influence of the input on the output nodes. Note that different CAMs can be generated from different output nodes, which can be interpreted as *an attention map to give a prediction as the target output*.

[Fig. 5 about here.]

From the CAM results, we found several characteristic patterns of wearable-sensor data for PD assessment Figure 5.

First, the parts that have no movement do not provide any informative evidence for PD assessment. Since no-motion patterns can be observed in any PD state, they are not salient for assessing the PD states. Therefore, CNNs do not pay attention to the parts with no-motion, rather focus on those parts that have movements. In that sense, our preprocessing to remove no-motion data is reasonable because they do not appear to contain any information for PD assessment.

Second, the *OFF* state is detected by the parts showing spikes or tremors which could be only a small part of the one-minute *OFF*-state window. Although slow or freezing movements are the most frequently observable patterns in the *OFF* state, they are not easy to distinguish from no-motion parts, thus, the CNN does not focus on them for detecting *OFF*-state. Instead, the CNN captures spiking or tremor patterns that are not commonly observed in *ON* and *DYS* states.

Third, smooth changes of the arm pose may indicate the *ON* state because it means that patients are able to control their arms. These movements are distinguishable from fluctuations during *DYS* state by the fact that the signals before and after the pose change are constant, indicating that patients have control of their arm movement. However, note that the *ON* state includes the largest variability of the movement patterns, thus, this finding is not sufficient to detect all possible instances of the *ON* state.

Lastly, continuously fluctuating signals are likely to represent *DYS* state. These fluctuations are distinguishable from tremors in that they present large and irregular movement patterns while tremors present regular oscillations with a frequency of 4-6 Hz. Also, they are distinguishable from voluntary movements because they do not contain stable pose holds. However, *DYS* and *ON* -state data are not easy to distinguish when patients are engaged in some activities. For example, some of our patients played dice games when they felt good (in *ON* state), however, the movements while shaking the dice are hard to distinguish from fluctuating movements in *DYS* state using only wearable-sensor data.

## 5 Conclusion

In this research, an approach for wearable-sensor-based PD assessment in free- living conditions is proposed using an ensemble of CNNs. To overcome the variability of the motion patterns over different patients, multiple CNNs trained with different patient groups are employed. Also, various aggregation strategies over individual models and time windows are investigated to improve the performance of PD assessment. As a result, accuracy of 63.05%, which is 8.89% higher than the baseline leave-one-out approach, is obtained for the classification of three PD states. The ensembles can also be used to estimate the reliability of predictions. For those patients who have at least 30% of top-20%-confidence predictions, an accuracy of 79.7 can be achieved by interpolating the high-confidence data. The signal features used by the CNN to distinguish between the motor state classes are visualized using CAM.

## Acknowledgement

The work of Mr. Um and Prof. Kulić was supported in part by Canada's Natural Sciences and Engineering Research Council and Ontario's Early Researcher Award. The work of Mrs. Lang and Dr. Endo was supported in part by the EU seventh framework programme FP7/2007-2013 within the ERC Starting Grant Control based on Human Models (con-humo), grant agreement no. 337654. Dr. Fietzek's position was supported by an unrestricted research grant from the Deutsche Stiftung Neurologie and the Deutsche Parkinson Vereinigung e.V.

## List of Figures

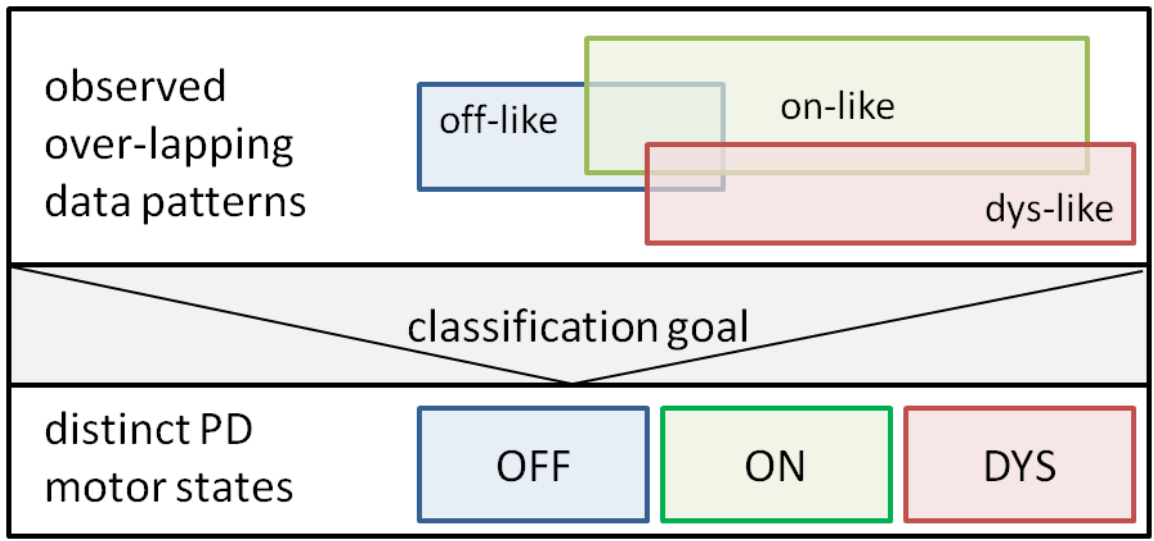

**Fig. 1** An abstract view of the three PD states: *OFF*, *ON*, and *DYS* states. The distributions of the observed signals have large overlaps with each other. In particular, due to the variability of voluntary movements, the *ON* state exhibits the largest variability and overlap with other states, thus, is the most challenging state to classify.

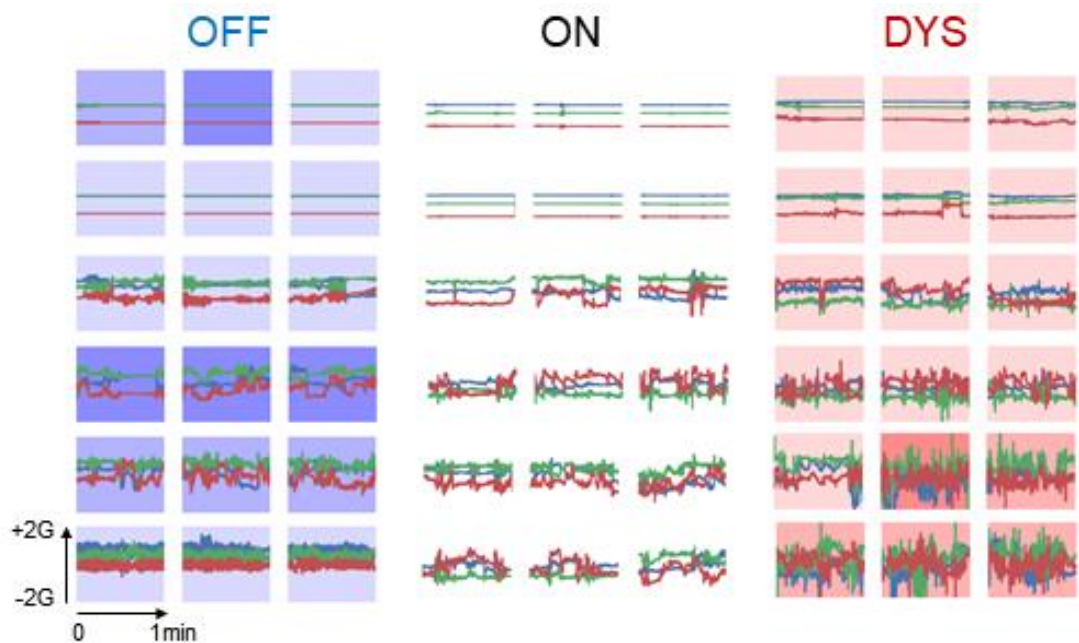

**Fig. 2** Examples of wearable sensor (accelerometer) data and their labels. RGB lines rep- resent XYZ components of the 1-min accelerometer signal and blue, white, red background colors represent *OFF*, *ON*, and *DYS* state labels, respectively. Note that a darker back- ground color indicates more severe symptoms. We can observe that a significant amount of data has similar waveforms but different labels. In other words, there exist large overlaps between classes, which makes our problem more challenging.

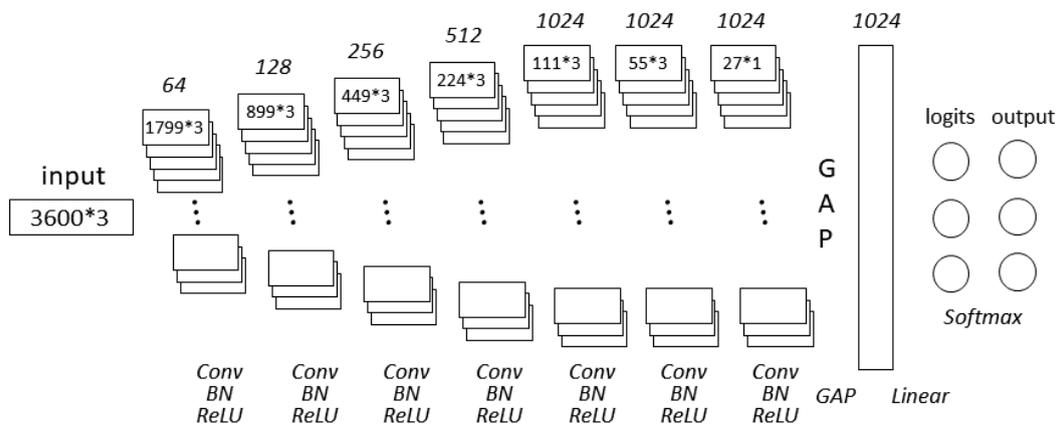

**Fig. 3** The 7-layer CNN used in this research. A global average pooling (GAP) layer is employed instead of fully-connected layers for reducing the number of parameters and enabling the visualization using CAM.

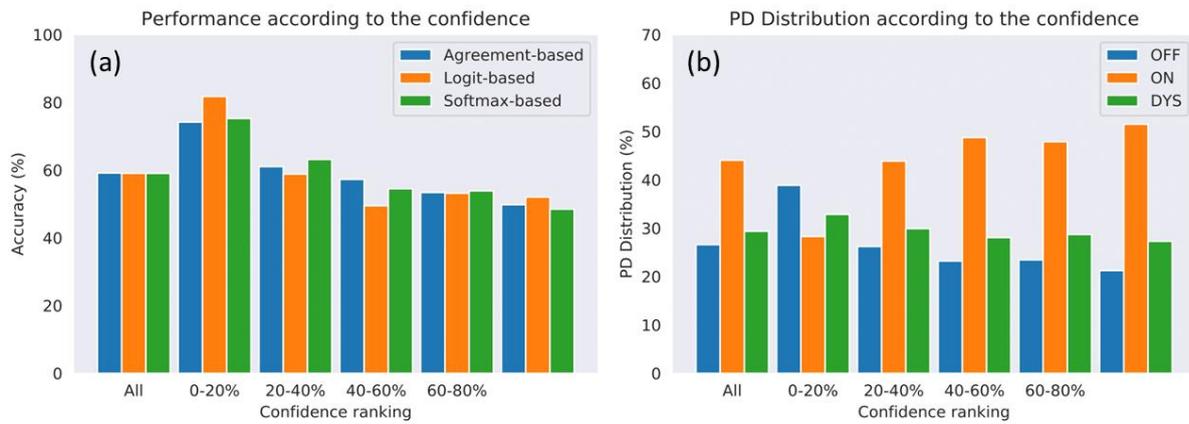

**Fig. 4 (a)** Predictions with high confidence show better classification accuracies than ones with low confidence. In the figure, 0-20% means the predictions which have the highest 20% confidence values. (b) *ON* state is the most challenging state to predict with high confidence.

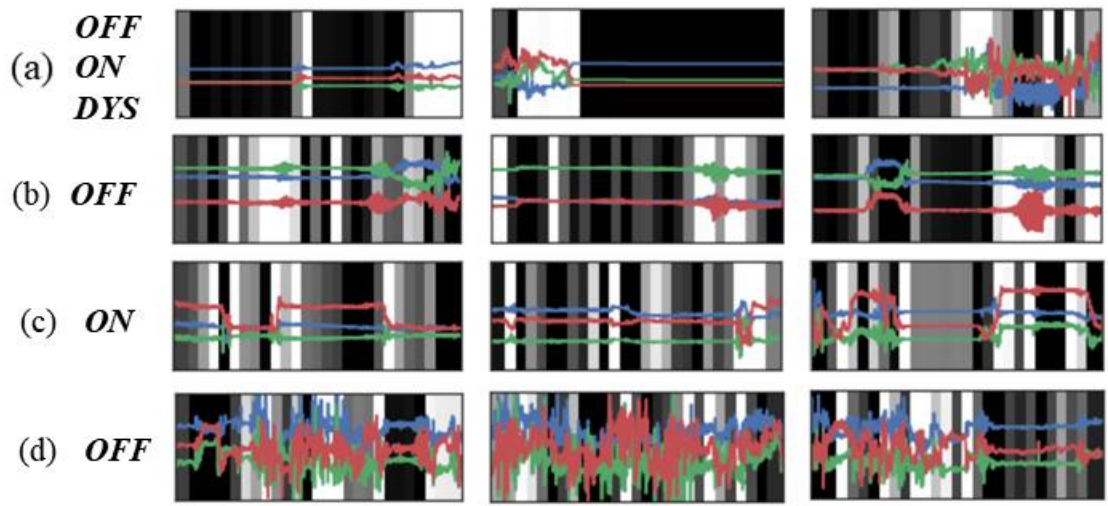

**Fig. 5** CAMs for one-minute wearable-sensor data. Light and dark background colors rep- resent the attended and the neglected parts, respectively. From the CAM results, we can notice that parts that have low signal magnitude are usually not attended by the CNN. Furthermore, the following characteristics of PD states can be observed: (a) Small amplitude oscillations appearing in the no-motion signal can be a clue for the *OFF* state. (b) Controlled hand-pose changes may indicate the *ON* state. (c) Continuously fluctuating movements are likely to indicate the *DYS* state.

## List of Tables

**Table 1**  PD distributions for 30 PD patients (min)

|       | 0   | 1   | 2   | 3   | 4   | 5   | 6   | 7   | 8   | 9   | 10  | 11  | 12  | 13  | 14  | 15  |
|-------|-----|-----|-----|-----|-----|-----|-----|-----|-----|-----|-----|-----|-----|-----|-----|-----|
| **OFF** | 385 | 81  | 97  | 29  | 136 | 87  | 62  | 59  | 163 | 0   | 0   | 99  | 220 | 160 | 42  | 12  |
| **ON**  | 14  | 142 | 27  | 531 | 119 | 314 | 224 | 19  | 178 | 412 | 0   | 273 | 197 | 199 | 355 | 83  |
| **DYS** | 0   | 239 | 144 | 0   | 182 | 123 | 166 | 356 | 218 | 13  | 162 | 3   | 0   | 0   | 36  | 242 |

|       | 16  | 17  | 18  | 19  | 20  | 21  | 22  | 23  | 24  | 25  | 26  | 27  | 28  | 29  | All    |
|-------|-----|-----|-----|-----|-----|-----|-----|-----|-----|-----|-----|-----|-----|-----|--------|
| **OFF** | 54  | 0   | 16  | 96  | 88  | 230 | 0   | 36  | 0   | 65  | 2   | 81  | 3   | 0   | **2303** |
| **ON**  | 71  | 0   | 27  | 0   | 285 | 0   | 8   | 83  | 39  | 52  | 106 | 40  | 10  | 5   | **3815** |
| **DYS** | 0   | 54  | 96  | 0   | 0   | 0   | 138 | 0   | 317 | 39  | 0   | 0   | 0   | 15  | **2543** |

**Table 2** Comparison of prediction accuracies between different learning approaches

|  | 0 | 1 | 2 | 3 | 4 | 5 | 6 | 7 | 8 | 9 | 10 | 11 | 12 | 13 | 14 | 15 |
|---|---|---|---|---|---|---|---|---|---|---|---|---|---|---|---|---|
| **CNN (n=15)** | 71.0 | 43.5 | 64.6 | 41.4 | 38.9 | 47.2 | 61.7 | 53.9 | 52.1 | 48.2 | 69.8 | 49.8 | 59.9 | 51.3 | 53.6 | 36.4 |
| **CNN (n=29)** | 83.4 | 44.1 | 76.1 | 47.0 | 39.6 | 57.1 | 62.8 | 45.2 | 51.7 | 32.0 | 40.1 | 62.4 | 62.8 | 57.1 | 65.4 | 28.8 |
| **Esb DropOut (n=29)** | 22.8 | 42.2 | 61.9 | 38.6 | 39.4 | 50.4 | 67.5 | 33.4 | 53.0 | 47.5 | 29.0 | 55.2 | 53.2 | 44.6 | 51.5 | 29.7 |
| **Esb DiffInit (n=15)** | 77.7 | 55.8 | 73.1 | 31.1 | 48.5 | 42.4 | 69.9 | 72.6 | 55.6 | 40.5 | 93.2 | 44.8 | 54.7 | 40.9 | 40.0 | 27.9 |
| **Esb DiffPat (n=15)** | 84.2 | 48.1 | 77.2 | 47.3 | 39.4 | 52.5 | 76.8 | 64.0 | 59.7 | 60.5 | 86.4 | 62.9 | 70.3 | 61.6 | 71.8 | 35.3 |

|  | 16 | 17 | 18 | 19 | 20 | 21 | 22 | 23 | 24 | 25 | 26 | 27 | 28 | 29 | All |
|---|---|---|---|---|---|---|---|---|---|---|---|---|---|---|---|
| **CNN (n=15)** | 30.2 | 50.6 | 49.7 | 30.4 | 44.3 | 74.3 | 26.3 | 42.6 | 58.4 | 46.4 | 32.7 | 69.1 | 52.5 | 71.3 | **50.97** |
| **CNN (n=29)** | 39.2 | 55.5 | 51.1 | 34.4 | 54.4 | 74.3 | 21.9 | 52.9 | 75.8 | 45.5 | 53.7 | 73.2 | 46.2 | 85.0 | **54.16** |
| **Esb DropOut (n=29)** | 22.4 | 22.2 | 38.1 | 18.8 | 42.9 | 18.3 | 28.1 | 49.6 | 35.7 | 43.6 | 48.1 | 43.1 | 53.8 | 50.0 | **43.19** |
| **Esb DiffInit (n=15)** | 32.0 | 51.9 | 53.2 | 13.5 | 60.6 | 81.3 | 27.4 | 49.6 | 57.3 | 47.4 | 58.3 | 73.2 | 61.5 | 75.0 | **52.74** |
| **Esb DiffPat (n=15)** | 21.6 | 61.1 | 66.2 | 8.3 | 50.9 | 80.4 | 23.3 | 39.5 | 72.8 | 48.7 | 32.4 | 78.9 | 69.2 | 85.0 | **59.14** |

**Table 3** PD assessment using 10-fold cross validation

| Fold | 0 | 1 | 2 | 3 | 4 | 5 | 6 | 7 | 8 | 9 | Avg.` |
|------|------|------|------|------|------|------|------|------|------|------|-------|
| Acc (%) | 62.4 | 65.8 | 64.3 | 64.7 | 68.4 | 63.6 | 58.8 | 67.6 | 64.9 | 66.3 | **64.67** |

**Table 4**  PD assessment with prediction smoothing

| Kernel Size (min) | 0 | 5 | 11 | 31 | 61 | 181 | Inf. |
|---|---|---|---|---|---|---|---|
| Accuracy (%) | 59.06 | 62.05 | 63.05 | 62.49 | 62.58 | 61.60 | 63.05 |

**Table 5**  PD distribution of the top 20% confidence data

|            | 0    | 1    | 2    | 3   | 4    | 5    | 6    | 7    | 8    | 9   | 10   | 11  | 12   | 13  | 14  | 15  |
|------------|------|------|------|-----|------|------|------|------|------|-----|------|-----|------|-----|-----|-----|
| **All**    | 399  | 462  | 268  | 560 | 437  | 524  | 452  | 434  | 559  | 425 | 162  | 375 | 417  | 359 | 433 | 337 |
| **TopConf**| 174  | 49   | 137  | 41  | 44   | 71   | 143  | 162  | 93   | 40  | 91   | 24  | 142  | 5   | 2   | 3   |
| **Top/All(%)** | 43.6 | 10.6 | 51.1 | 7.3 | 10.1 | 13.5 | 31.6 | 37.3 | 16.6 | 9.4 | 56.2 | 6.4 | 34.1 | 1.4 | 0.5 | 0.9 |

|            | 16  | 17   | 18   | 19  | 20  | 21   | 22  | 23   | 24   | 25   | 26   | 27   | 28  | 29   | All     |
|------------|-----|------|------|-----|-----|------|-----|------|------|------|------|------|-----|------|---------|
| **All**    | 125 | 54   | 139  | 96  | 373 | 230  | 146 | 119  | 356  | 156  | 108  | 123  | 13  | 20   | **8661**|
| **TopConf**| 3   | 15   | 32   | 3   | 13  | 152  | 8   | 13   | 154  | 28   | 27   | 55   | 0   | 8    | **1732**|
| **Top/All(%)** | 2.4 | 27.8 | 23.0 | 3.1 | 3.5 | 66.1 | 5.5 | 10.9 | 43.3 | 17.9 | 25.0 | 44.7 | 0.0 | 40.0 | **20.0**|